\documentclass[lettersize, journal]{IEEEtran}

\usepackage[utf8]{inputenc}       
\usepackage[T1]{fontenc}          
\usepackage{microtype}            
\usepackage{xcolor}               

\usepackage{amsmath}              
\usepackage{amsfonts}             
\usepackage{nicefrac}             

\usepackage{graphicx}             
\usepackage{float}                
\usepackage{booktabs}             
\usepackage{multirow}             
\newfloat{figtab}{htb}{fgtb}      
\makeatletter
\newcommand\figcaption{\def\@captype{figure}\caption}
\newcommand\tabcaption{\def\@captype{table}\caption}
\makeatother

\usepackage{algorithm}            
\usepackage{algpseudocode}        

\usepackage{url}                  
\usepackage{hyperref}            
\usepackage[numbers, sort&compress]{natbib} 

\usepackage{lineno}
\usepackage{etoolbox}
\makeatletter
\patchcmd{\@twocolumnfalse}{\setpagewiselinenumbers}{\setpagewiselinenumbers\linenumbers}
{\setpagewiselinenumbers\linenumbers}{}
\makeatother

\def\BibTeX{{\rm B\kern-.05em{\sc i\kern-.025em b}\kern-.08em
  T\kern-.1667em\lower.7ex\hbox{E}\kern-.125emX}}

\newcommand{\revised}[1]{{\color{black}#1}}

\title{iFuzz-Meta: An Interpretable Fuzzy Learning Framework Bridging Top-Down and Bottom-Up Knowledge Integration}

\author{
        Xiaowei Jiang\textsuperscript{*}, Daniel Leong, Beining Cao, Nan Zhou, Yingtao Ren, Yu-Cheng Chang, Thomas Do, and Chin-Teng Lin, \textit{Fellow, IEEE}
    
    \thanks{All authors are with the Computational Intelligence and Brain Computer Interface Lab, Australian AI Institute, School of Computer Science, Faculty of Engineering and Information Technology, University of Technology Sydney.}
    
    \thanks{This work was supported in part by the Australian Research Council (ARC) under discovery grant DP250103612 and DP260101395, ARC Research Hub for Human-Robot Teaming for Sustainable and Resilient Construction (ITRH) grant IH240100016, and Australian National Health and Medical Research Council (NHMRC) Ideas Grant APP2021183.}
    \thanks{\textsuperscript{*}Corresponding author: Xiaowei Jiang (email: xiaowei.jiang-1@student.uts.edu.au)}
}

\begin{document}

\maketitle

\begin{abstract}
Interpretable representation learning remains a key challenge in modern neural computation, particularly when models are expected not only to perform but also to explain their reasoning. This paper introduces iFuzz-Meta, an interpretable fuzzy rule-based learning framework that preserves human-understandable reasoning structures within modern neural architectures. Each fuzzy rule corresponds to a semantic and spatial prototype defined in the original feature space, enabling transparent inference and direct interpretability. Meta-learning is employed as an analytical paradigm to examine how these interpretable rules reorganize across tasks and domains, providing a principled means to link algorithmic adaptation with cognitive representation. A knowledge-guided regularization mechanism further enables a \emph{top-down–bottom-up} integration, in which theoretical priors act as soft inductive biases while data-driven learning refines and extends them. This dual process ensures that adaptation proceeds along semantically and physiologically meaningful trajectories, rather than arbitrary parameter shifts. Evaluations demonstrate that iFuzz-Meta achieves interpretable reasoning and stable cross-domain generalization, establishing a \revised{potential general} pathway toward explainable and knowledge-aware fuzzy systems.
\end{abstract}

\begin{IEEEkeywords}
Fuzzy neural networks, interpretable representation learning, meta-learning, domain adaptation.
\end{IEEEkeywords}

\section{Introduction}

\begin{figure}
    \centering
    \includegraphics[width=1\linewidth]{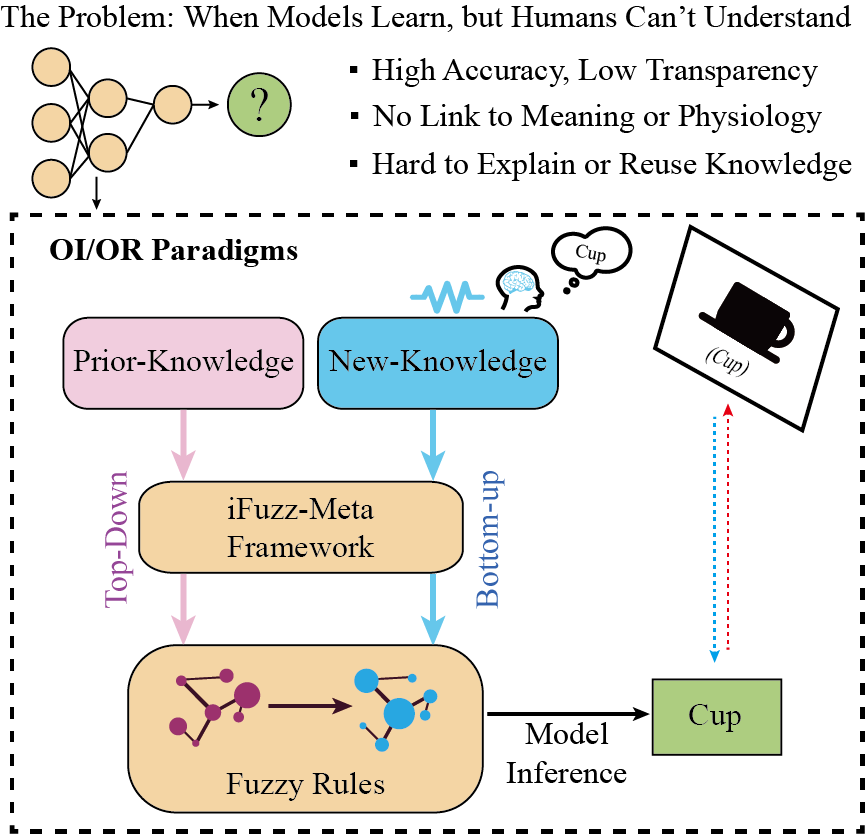}
    \caption{\textbf{Conceptual overview of the proposed iFuzz-Meta framework.} The framework addresses the opacity of conventional deep models by grounding learning in human-understandable fuzzy rule centers. Top-down prior knowledge and bottom-up data-driven adaptation jointly shape the evolution of these rule centers, enabling interpretable, stable, and knowledge-guided reasoning. This interaction preserves transparency while supporting meaningful model adaptation across domains.}
    \label{fig:placeholder}
\end{figure}

\IEEEPARstart{I}{nterpretable} representation learning remains a central challenge in modern neural computation. As deep learning models continue to achieve remarkable predictive accuracy across domains, their inherent opacity poses fundamental obstacles to scientific understanding, knowledge transfer, and trustworthiness~\cite{rajpuraExplainableArtificialIntelligence2024}. This tension is particularly evident in neurocomputational systems, where the aim is not only to decode neural activity but also to reveal the mechanisms by which cognition and intention emerge from brain dynamics~\cite{lin2020direct}. In such settings, the ability to link algorithmic representations to interpretable neurophysiological phenomena is indispensable.  

Conventional neural networks~\cite{lawhernEEGNetCompactConvolutional2018}, LSTM~\cite{shibuExplainableArtificialIntelligence2023} or Transformer-based models~\cite{eegformer}, though powerful, achieve their performance by learning highly entangled internal representations that lack semantic transparency and explainability. Once trained, the mapping from input signals to latent features becomes difficult to trace or relate to domain knowledge. When applied to neural decoding tasks, this limitation prevents the incorporation of well-established neuro-scientific priors such as frequency-band specificity or spatial modularity, and renders the learned features opaque to human understanding. Some approaches attempt to address this issue through multi-bank architectures, such as filter-bank based SSVEPFormers~\cite{CHEN2023521}, that explicitly partition inputs according to predefined frequency bands. However, such designs rely heavily on the correctness of prior knowledge. When the underlying theory is imperfect or incomplete~\cite{varoquaux2021ai,10849810}, these hard-coded priors may introduce structural bias, restrict the model’s self-organizing capacity, and lead to suboptimal generalization despite increased model complexity and computational cost. 

\begin{figure*}[t!]
    \centering
    \includegraphics[width=1\linewidth]{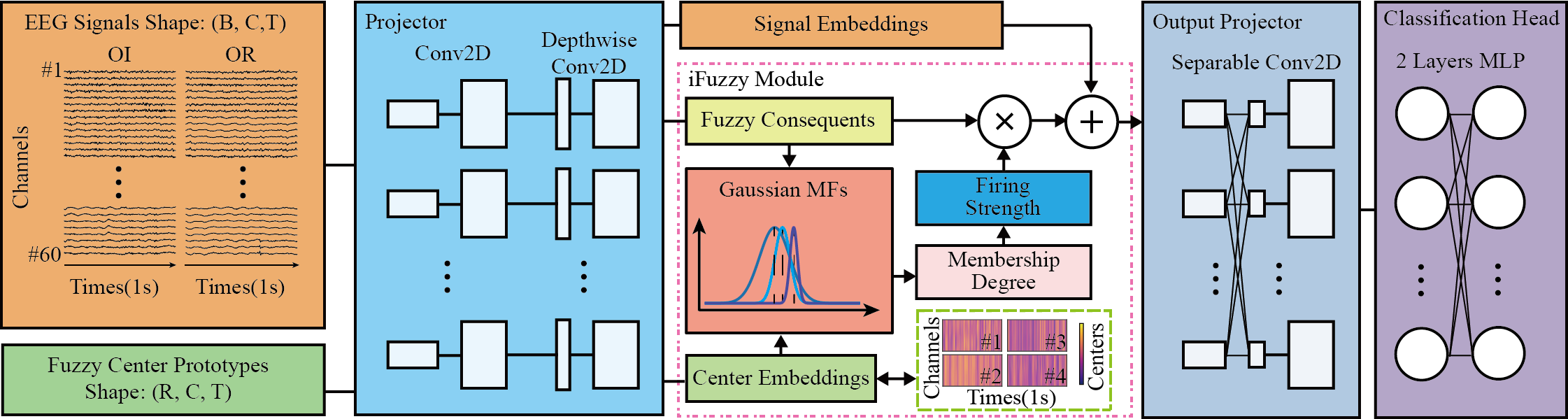}
    \caption{\revised{\textbf{Model Architecture.} Fuzzy center prototypes are first defined in the raw neurophysiological space ($B$: batch size. $C$: number of channels. $T$: number of timestamps). Both EEG inputs and fuzzy centers ($R$: number of rules) are projected into a shared latent space using a structured convolutional encoder. The illustrated configuration adopts a Temporal-Spatial Pattern Analysis (TSpPA) design. Fuzzy embeddings are computed using Gaussian membership functions within the iFuzzy module, and the resulting rule activations introduce interpretable knowledge to guide the modulation of feature representations. A two-layer MLP serves as the final classification head.}}
    \label{fig:MainModelStructure}
\end{figure*}

In practice, researchers are forced to rely on post-hoc visualization or correlation analysis to interpret model outputs, such as SHAP \cite{shap2017} and gradient-based approaches like Grad-CAM \cite{8237336}. These approaches neither \revised{guarantee} causal validity nor offers insight into what the model has truly learned. Moreover, such explanations lack task-level reusability and cannot serve as transferable knowledge for improving model generalization \cite{Kamath2021,han2022which,Janzing2020}. 
Although attention mechanisms in transformer-based BCI models have shown potential for enhancing interpretability, recent work indicates that their explanatory power remains limited without explicit alignment to neural or cognitive priors \cite{wang2025integrating}. \revised{In particular, the key, query, and value representations in attention mechanisms are learned in an unconstrained latent space and lack direct correspondence to physically meaningful EEG patterns.} Unlike attention, fuzzy reasoning provides discrete, rule-based prototypes that offer clear semantic pathways. This contrast motivates incorporating fuzzy rules to restore transparent, concept-level interpretability in deep neural models. 

Fuzzy logic systems offer a principled alternative by encoding human-understandable reasoning rules in the form of linguistic terms and membership functions. Unlike purely numerical optimization, fuzzy inference provides a transparent mapping between features and decisions through the explicit definition of fuzzy sets and rules\revised{~\cite{lin1996neural,931548,10536605,9961931,8071166}}. Each rule corresponds to a local model whose parameters have interpretable meanings such as prototype centers or feature importances. This property enables fuzzy systems to bridge data-driven learning and expert knowledge, making them particularly attractive for domains, such as brain-computer interfaces (BCIs)~\cite{cao2025emdfuzzy,jiang2025interpretable, do2020estimating, hesam2020evaluation, nguyen2025edge}, where interpretability and physiological grounding are essential.

Recent advances in fuzzy-neural modeling further demonstrate that rule-based reasoning and differentiable representation learning can be seamlessly integrated. Fuzzy Neural Networks (FNNs), typically grounded in Takagi-Sugeno-Kang (TSK) inference systems~\cite{shihabudheen2018recent,luFuzzyMachineLearning2024}, combine the expressiveness of deep architectures with the transparency of fuzzy reasoning, and can be efficiently optimized through gradient-based learning~\cite{10183374,106218}. Such frameworks have been successfully applied across a variety of cognitive and affective computing tasks, including social cognitive tasks~\cite{10849810}, affective computing~\cite{jiang2025interpretable}, physical cognitive estimation~\cite{do2020estimating, do2021retrosplenial, lin2022effects}, and steady-state visual evoked potential (SSVEP) classification in BCIs~\cite{cao2025emdfuzzy,jiang2024ifuzzytl}. In transfer learning contexts, fuzzy rule-based architectures such as iFuzzyTL~\cite{jiang2024ifuzzytl} demonstrate how shared rule centers and adaptive membership functions can retain and adapt knowledge in an interpretable manner across subjects or domains. Collectively, these developments highlight that fuzzy logic not only enhances interpretability but also provides a systematic and mathematically grounded framework for transparent, knowledge-guided adaptation in complex neural and cognitive systems.

From an algorithmic perspective, conventional fuzzy systems still face fundamental challenges in achieving genuine interpretability under modern learning settings. In most existing implementations, fuzzy centers or membership functions are computed within high-dimensional embedding spaces produced by neural networks or other representation learners. Although these embeddings allow gradient-based optimization and yield numerically stable rule adaptation, the resulting rule centers no longer correspond to explicit or human-understandable concepts. Consequently, the interpretability becomes representational rather than semantic, that the model can indicate \emph{where} a rule changes in the latent space, but not \emph{why} it changes in relation to the underlying mechanism. This limitation is particularly pronounced when applied to neural data, where shifts in rule centers cannot be readily linked to meaningful neurophysiological processes such as microstate transitions or cognitive-state dynamics. Thus, while fuzzy systems are theoretically interpretable, their modern implementations often lose explanatory transparency when fused with embedding-based learning\revised{~\cite{jiang2024ifuzzytl, 10849810,cao2025emdfuzzy,jiang2025interpretable}}.

\begin{figure*}[!t]
    \centering
    \includegraphics[width=1\linewidth]{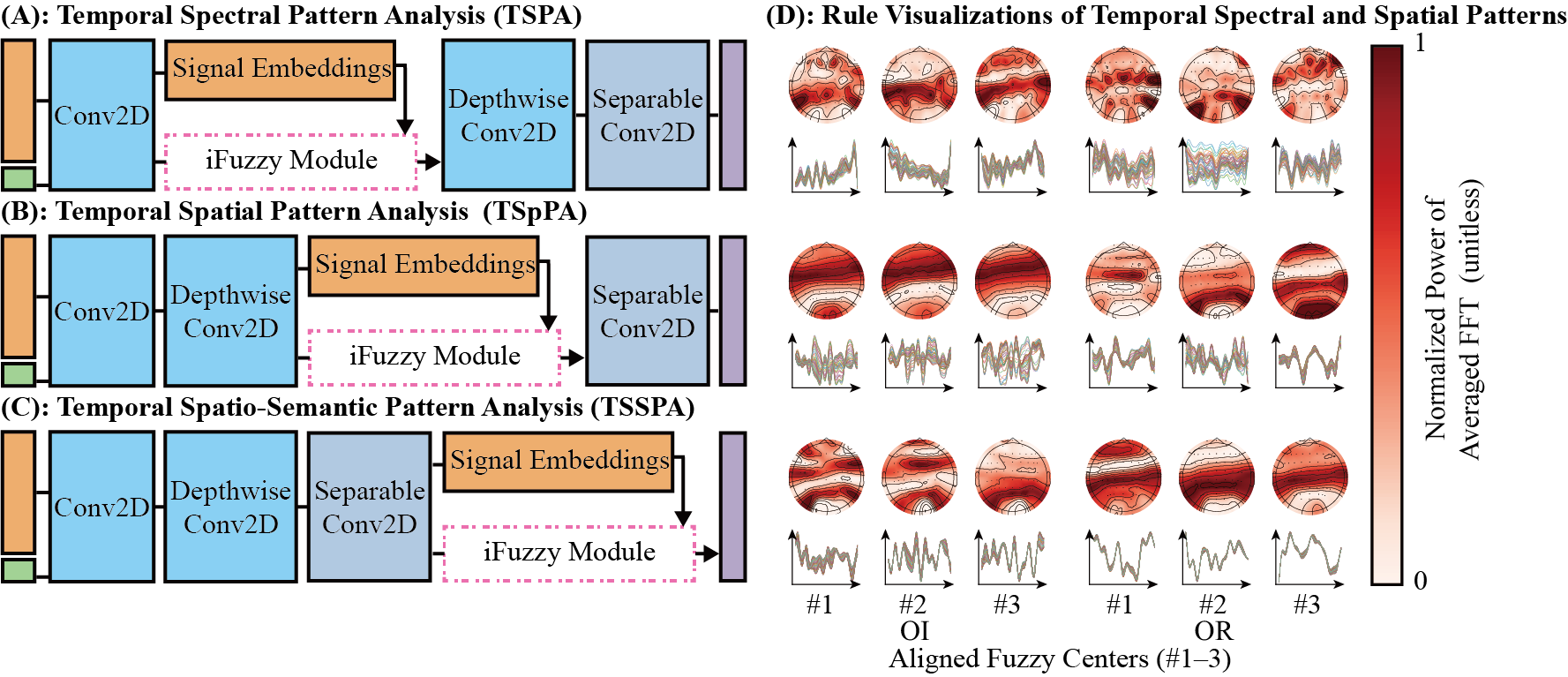}
    \caption{
        \textbf{(A) Temporal Spectral Pattern Analysis (TSPA) framework.} Designed to extract temporal dynamics from EEG signals using early temporal convolution layers. 
        \textbf{(B) Temporal Spatial Pattern Analysis (TSpPA) framework.} Enhances spatial information extraction by introducing spatial filters after initial temporal processing. 
        \textbf{(C) Temporal Spatio-Semantic Pattern Analysis (TSSPA) framework.} Integrates both temporal and spatial features to capture high-level semantic representations.
        \textbf{(D) Visualization of fuzzy center prototypes.} Topographic maps show the average spectral power in the slow frequency band (0.5-8~Hz) for three representative fuzzy centers after meta-learning and fine-tuning under each architecture.
    }

    \label{fig:model_structures}
\end{figure*}

\revised{To address these limitations, we propose iFuzz-Meta, an interpretable fuzzy rule-based framework designed to reveal how human-understandable reasoning structures evolve across cognitive tasks. In iFuzz-Meta, each fuzzy rule corresponds to a spatially grounded and semantically meaningful neural prototype. Specifically, a human-understandable prototype refers to a fuzzy rule center defined in the original EEG signal space, such that each prototype corresponds to a spatiotemporal pattern that can be interpreted in terms of waveform morphology, spectral content, and spatial distribution. As a result, interpretability arises intrinsically from the fuzzy reasoning mechanism, since each rule center represents a meaningful signal pattern rather than an abstract latent embedding or a post-hoc attribution map over model activations. This design further enables inspection and visualization of the learned prototypes, allowing researchers to examine how neural response patterns are captured and used during the decision process.

A critical distinction between the proposed framework and established fuzzy neural networks lies in the geometric and conceptual mapping of the prototype definition space. Traditional methods typically define fuzzy prototypes in latent space~\cite{jiang2024ifuzzytl, 10849810}, which improves optimization but weakens semantic interpretability. In contrast, our framework defines fuzzy center prototypes in raw space and performs fuzzy inference in latent space, thereby combining interpretability with representation learning capability.}

\begin{figure}
    \centering
    \includegraphics[width=1\linewidth]{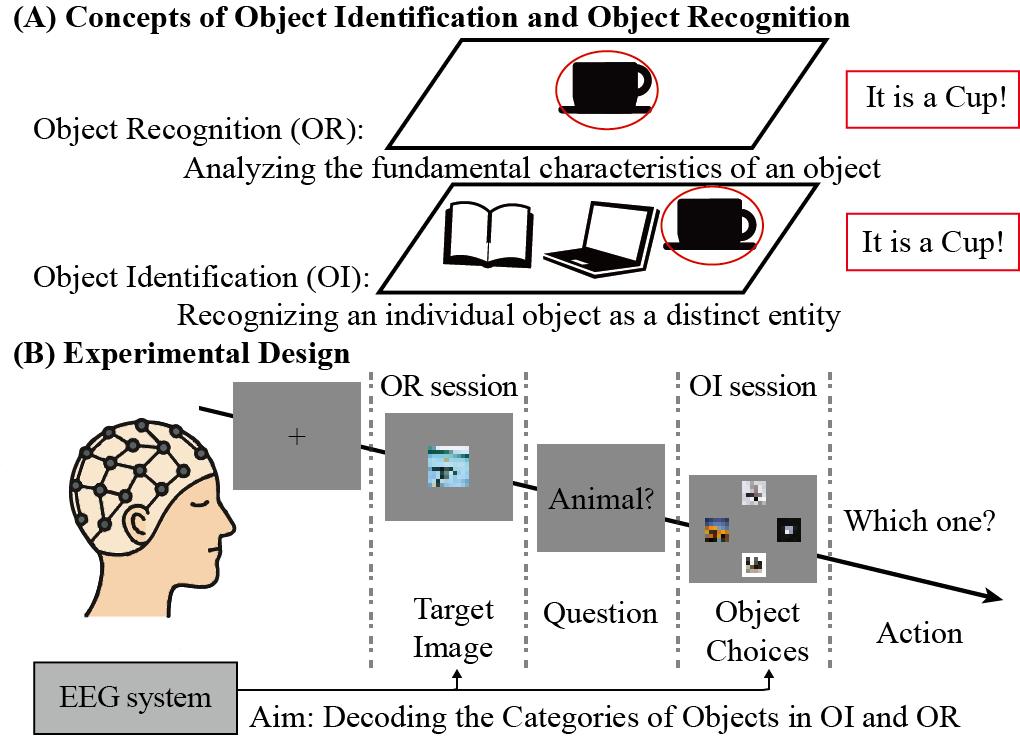}
    \caption{\revised{\textbf{(A) Task Definitions.} Object Recognition (OR) refers to decoding the underlying object category, while Object Identification (OI) refers to recognizing a specific object instance \cite{10570151}. An example is shown where both tasks are applied to a visual stimulus depicting a cup.
    \textbf{(B) Experimental Design.} A schematic of the experimental setup used to record EEG signals for both OI and OR paradigms. In the OR paradigm, participants are asked to report the category of a target object. Subsequently, in the OI paradigm, they are required to identify which of the four objects (up, down, left, right) belongs to the same category as the target.}}
    \label{fig:exp_design}
\end{figure}

\revised{
Meta-learning is introduced here as a cross-subject analytical paradigm for few-shot adaptation under a leave-one-subject-out (LOSO) protocol, rather than as a mechanism for class transfer or class-increment learning.
}
Specifically, we employ the paradigms of \emph{object recognition} (OR) and \emph{object identification} (OI)~\cite{leong2023ventral,10570151}, which allow us to investigate how semantic-level processing (OR) transitions into instance-specific individuation (OI) in both neural and rule space. OR focuses on categorical recognition (e.g., identifying an image as a “cup”), whereas OI requires individuating a specific exemplar among visually similar alternatives (e.g., selecting “this cup” among multiple objects), as illustrated in \revised{Fig.~\ref{fig:exp_design}(A)}. These two sub-tasks form a cognitively meaningful hierarchy that mirrors the transition from perception to intention formation. Embedding this structure within a meta-learning formulation allows us to quantify and visualize how fuzzy rule centers shift along this semantic-intentional continuum, thereby linking interpretable algorithmic representations to neuro-cognitive processes. In this setting, iFuzz-Meta serves as both a transparent modeling framework and a scientific tool to study human-intention-related representational changes in a human-understandable manner. \revised{In our setting, OR and OI are used as cognitively grounded meta-tasks to distill intention related neural regularities that are shared across subjects. The meta-learning objective is to learn transferable fuzzy rule priors and an initialization from these intention paradigms, rather than only to transfer a classifier from 2 classes to 4 classes.}

Beyond its interpretability, the proposed framework also enables the principled integration of domain knowledge into the learning process, even when the underlying theory is imperfect or incomplete. To achieve this, we introduce a frequency-regularized center loss that explicitly incorporates neuro-scientific priors regarding the spectral organization of cognitive processes. Empirical evidence indicates that distinct EEG frequency bands correspond to different cognitive functions~\cite{gong2024cortical}. Building on this understanding, the loss function constrains each fuzzy prototype to maintain a physiologically meaningful spectral profile, ensuring that the learned rule centers correspond not only to discriminative but also neuro-cognitively plausible features.

Rather than serving as a purely data-driven regularizer, the proposed mechanism embodies a \emph{top-down–bottom-up} integration. Neuroscientific \revised{prior theory and knowledge} provide soft inductive biases from the top down, while the model’s learning dynamics iteratively refine and reinterpret these biases from the bottom up, forming a reciprocal loop between theory and data. 
\revised{Concretely, this bidirectional interaction is instantiated at the optimization level, where prior-informed regularization and task-driven losses jointly shape the gradient updates of the same fuzzy prototypes.}
This dual process reflects the view that effective computational frameworks should unify theoretical and statistical perspectives: top-down computational theories offer structured inductive constraints, whereas bottom-up empirical inference grounds and revises those theories through data-driven evidence~\cite{Linderman2017,Griffiths2010,Mlynarski2019}. This bidirectional integration establishes a principled pathway for embedding theoretical knowledge into adaptive fuzzy systems. Within the OR and OI paradigm, such coupling enables the model to incorporate domain knowledge without rigidly imposing it, thereby enhancing both the utilization and the understanding of imperfect theory.

\noindent\textbf{In summary, the contributions of this study are threefold:}

\noindent\textbf{(1): Fuzzy Interpretability with Human-Understandable Prototypes.}  
    We achieve intuitive interpretability by preserving fuzzy rule centers in the original feature space, allowing their spatial distributions to be directly visualized and semantically interpreted.

\noindent\textbf{(2): Cognitive-Level Meta-Analytical Framework.}  
    By reframing meta-learning as an analytical paradigm, our framework enables explicit observation of how fuzzy rule centers evolve during domain adaptation. This interpretability allows researchers to monitor the direction and structure of adaptation, ensuring that the model learns along neurocognitively meaningful trajectories rather than arbitrary or unstable parameter shifts. Such transparency is critical for validating whether domain adaptation preserves the intended semantic relationships across tasks and subjects.
    
\begin{figure}[t!]
    \centering
    \includegraphics[width=1\linewidth]{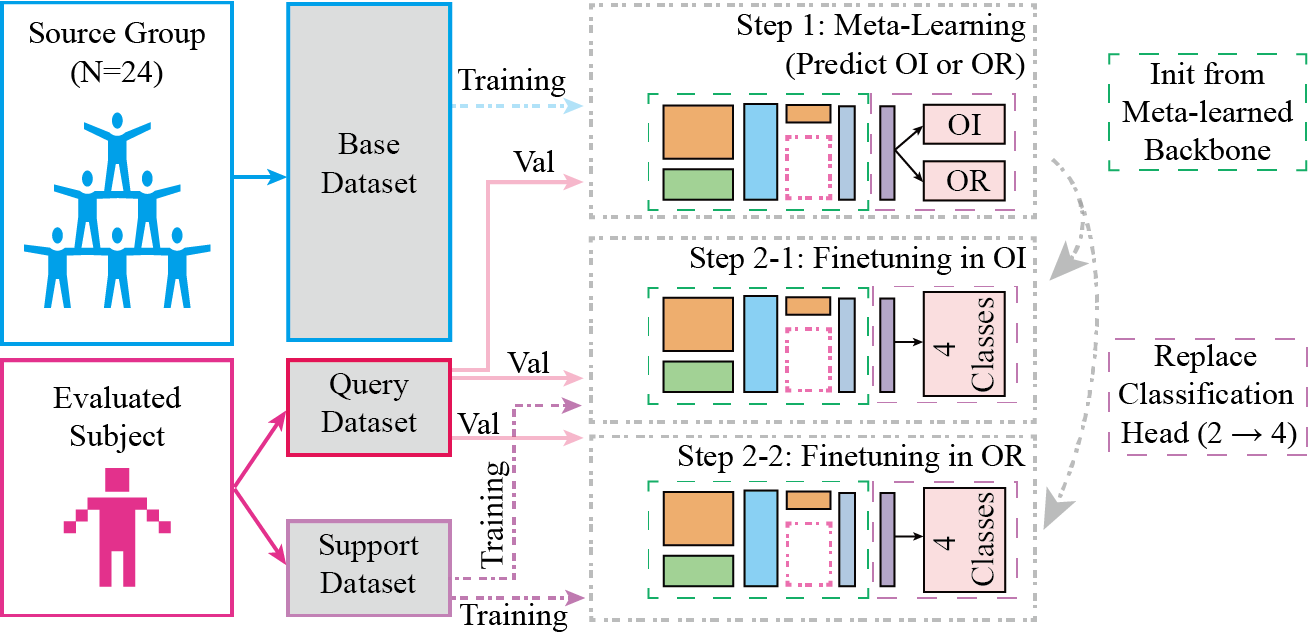}
    \caption{\revised{\textbf{Meta-Learning Paradigm.} The training process is structured into two stages using three datasets: a base set from the source group and query/support sets from the target subject. Step~1 (meta-learning) constructs \emph{cognitively structured binary meta-tasks} based on Object Identification (OI) and Object Recognition (OR) to learn transferable, subject-generalizable priors. Step~2 (fine-tuning) performs standard few-shot adaptation on the target subject, where the classification head is adapted for four-class object recognition as the final evaluation task.}}
    \label{fig:dataflow}
\end{figure}

\noindent\textbf{(3): Knowledge-Guided Regularization Framework.}  
    We establish a \revised{potential general pathway} for incorporating domain knowledge into fuzzy learning by introducing knowledge-guided regularization as a soft inductive bias. This framework enables top-down theoretical constraints to interact with bottom-up data-driven adaptation, guiding the model toward physiologically and semantically meaningful representations while enhancing stability, transferability, and interpretability across tasks and domains.

The remainder of this paper is structured as follows. 
Section~\ref{sec:methods} introduces the proposed iFuzz-Meta framework, including the cognitive task formulation, encoder architectures, fuzzy-rule representation, and the knowledge-guided regularization mechanism. 
Section~\ref{sec:results} presents comprehensive evaluations under few-shot cross-subject settings, ablation studies on rule structures and regularization, and analyses of the interpretability and neurophysiological meaning of the learned fuzzy prototypes. 
Section~\ref{sec:discussion} discusses the cognitive implications of the OR--OI hierarchy, the stability of rule prototypes, and the balance between interpretability and model capacity. 
Finally, Section~\ref{sec:conclusion} concludes the study and outlines future research directions.

\section{Methods}
\label{sec:methods}

\subsection{Problem Formulation and Meta-Task Construction}

\subsubsection{Data Representation and Task Setup}
The dataset description and the experimental design details (\revised{Fig.~\ref{fig:exp_design}(B)}) are shown in Supplemental Material Section I-A.
EEG recordings from the two paradigms,OR and OI,are organized as trial-wise multichannel time series \( x \in \mathbb{R}^{C \times T} \), where \( C = 60 \) denotes the number of EEG channels and \( T = 1000 \) represents the number of time points sampled over a 1-second interval following stimulus onset. Each trial is annotated with a label \( y \in \{1, 2, 3, 4\} \), indicating one of four object classes.

For each subject \( s \in \{1, \dots, N\} \), we denote the dataset as \( \mathcal{D}_s = \{(x_i^{(s)}, y_i^{(s)})\}_{i=1}^{n_s} \). The objective is to learn a classifier \( f_\theta : \mathbb{R}^{C \times T} \rightarrow \{1,2,3,4\} \), parameterized by \( \theta \), that generalizes to a previously unseen target subject \( t \), given access only to source domain data from the remaining \( N - 1 \) subjects.

As shown in \revised{Fig.~\ref{fig:dataflow}}, we adopt a \revised{LOSO} few-shot meta-learning protocol to evaluate cross-subject generalization. In each evaluation episode, the data is partitioned into:

\begin{itemize}
    \item \textbf{Base set} \( \mathcal{D}_{\text{base}} = \bigcup_{s \ne t} \mathcal{D}_s \), used for meta-training;
    \item \textbf{Support set} \( \mathcal{D}_{\text{support}}^{(t)} \subset \mathcal{D}_t \), a small labeled subset from the target subject for adaptation;
    \item \textbf{Query set} \( \mathcal{D}_{\text{query}}^{(t)} = \mathcal{D}_t \setminus \mathcal{D}_{\text{support}}^{(t)} \), used for evaluation.
\end{itemize}

The underlying assumption is that base subjects provide access to shared cognitive and neurophysiological priors that can be leveraged to construct a transferable feature space. However, the target subject often introduces substantial variability \( \delta^{(t)} \), which may be systematic (e.g., head geometry, electrode impedance) or stochastic (e.g., arousal state, attention) \cite{9714736,8914469}. Therefore, given latent feature encoder \( p_0: \mathbb{R}^{C \times T} \rightarrow \mathbb{R}^D \), the observed EEG from the target subject can be approximated as:
\begin{equation}
    z^{(t)} = p_0(x^{(t)}) = z^{\text{shared}} + \delta^{(t)}.
\end{equation}
Here, \( z^{\text{shared}} \) represents generalizable structure learned from base subjects, while \( \delta^{(t)} \) captures subject-specific deviations. Few-shot adaptation aims to mitigate the impact of \( \delta^{(t)} \) by fine-tuning fuzzy centers and decision layers based on a limited support set, using only \( |\mathcal{D}_{\text{support}}^{(t)}| \ll |\mathcal{D}_t| \), enabling effective generalization.

\begin{figure}[!t]
    \centering
    \includegraphics[width=1\linewidth]{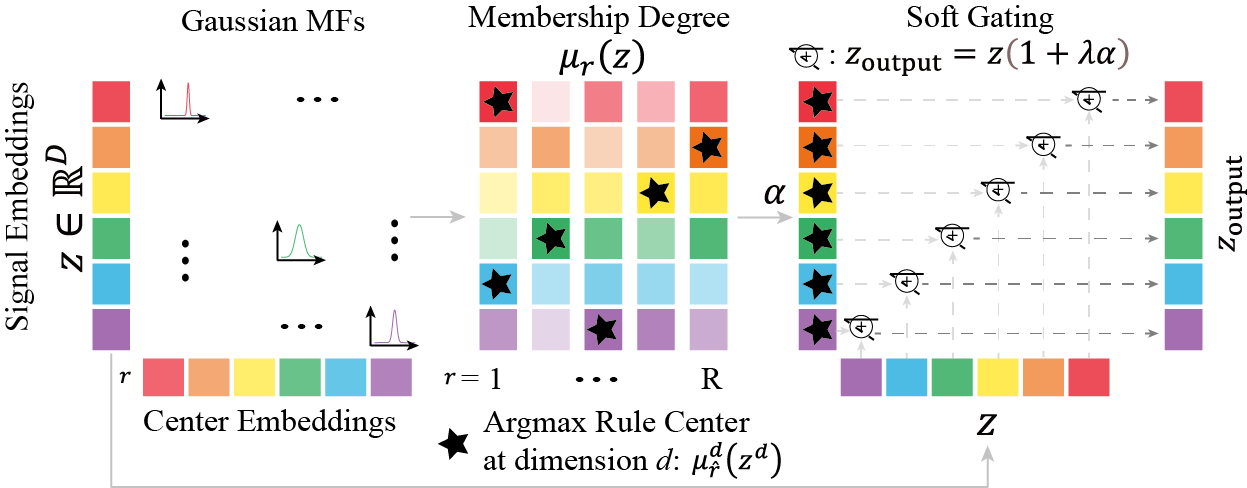}
    \caption{Illustration of the iFuzzy Network Module. Given the input signal embedding \( z \in \mathbb{R}^D \), each rule \( r \) is associated with a center embedding \(c_r\) and a Gaussian MF. Membership degrees \( \mu_r(z) \) are computed per dimension \(d\), with the maximum activation dimension marked by a star at rule \(\hat{r}\). These values are aggregated into an attention vector \( \alpha \), which softly modulates the input to produce the gated output, following eq. (\ref{eq:modulation}).}
    \label{fig:fuzzy_detail}
\end{figure}

\subsubsection{From Meta-Tasks to Class-Level Generalization}

To construct structured learning signals, we define meta-tasks based on the OI and OR paradigms. These tasks differ in their cognitive abstraction levels: OI requires distinguishing between visually similar object instances, whereas OR focuses on abstract semantic categorization. We denote the task distribution as \( \mathcal{T} = \{\tau_k\} \), where each task \( \tau \in \mathcal{T} \) contains a support set \( \mathcal{D}_{\text{support}}^\tau \) and a query set \( \mathcal{D}_{\text{query}}^\tau \). This setup encourages the model to learn feature hierarchies that disentangle instance-level specificity from category-level invariance, thereby enabling semantic generalization across subjects and paradigms.

Learning is formulated as a bi-level meta-optimization problem comprising inner-loop and outer-loop objectives:

\paragraph{Inner-loop objective.} For a sampled task \( \tau \sim \mathcal{T} \), model parameters \( \theta \) are updated using task-specific loss \( \mathcal{L}_{\text{task}}^\tau \) via one-step gradient descent:
\begin{equation}
\theta' = \theta - \eta \nabla_\theta \mathcal{L}_{\text{task}}^\tau(\theta),
\end{equation}
where \revised{\( \eta = 1.5e-4\)} is the inner-loop learning rate.

\paragraph{Outer-loop objective.} The updated parameters \( \theta' \) are evaluated on the corresponding query set, yielding the meta-objective:
\begin{equation}
\min_{\theta, p_0} \ \mathbb{E}_{\tau \sim \mathcal{T}} \left[ 
    \mathcal{L}_{\text{meta}}^\tau(\theta') 
\right].
\end{equation}
This formulation encourages the encoder \( p_0 \) to extract class-discriminative and subject-invariant representations, while allowing rapid subject-specific adaptation through the task-specific parameter update \( \theta \mapsto \theta' \). The meta-learner thus aims to separate transferable priors from subject-dependent variability, enabling robust cross-subject generalization in low-resource EEG settings. \revised{The learning rate \( \eta\) is also $1.5e-4$}.

\subsection{Model Architecture Overview}
\label{sec:model_arch}
\revised{Fig.~\ref{fig:MainModelStructure}} show the detailed structures of our proposed iFuzz-Meta framework. The encoder \( p_0: \mathbb{R}^{D_{\text{raw}}} \rightarrow \mathbb{R}^{D} \) serves to project raw EEG signals into a latent feature space conducive to fuzzy reasoning. In this work, following the baseline established in \cite{leong2023ventral}, we instantiate the encoder \( p_0 \) using EEGNet \cite{lawhernEEGNetCompactConvolutional2018}, due to its proven effectiveness in neural signal decoding and its compact, interpretable architecture. To explore different levels of neural representation, we implement three distinct frameworks of \( p_0 \), each corresponding to a specific stage within EEGNet, as shown in Fig.  \ref{fig:model_structures}(A, B, and C):
\textbf{TSPA} (Temporal-Spectral Pattern Analysis): utilizes the output from EEGNet’s initial temporal convolution layer, capturing fine-grained spectral dynamics across time.
\textbf{TSpPA} (Temporal-Spatial Pattern Analysis): extracts features after the depthwise spatial convolution, enabling the model to leverage spatially resolved, frequency-specific patterns for fuzzy inference.
\textbf{TSSPA} (Temporal-Spatial-Semantic Pattern Analysis): adopts the final latent representation from EEGNet, integrating temporal and spatial features into a high-level semantic embedding.

These variants capture progressively abstract features, enabling flexible integration with the fuzzy module across different processing stages.

\subsection{Fuzzy Inference System (FIS) Foundations}

\subsubsection{Mamdani-style Inference for Neural Modulation}

\revised{Unlike attention mechanisms that aim to aggregate information from multiple latent components, the objective of the proposed fuzzy inference system is to attribute each input to an explicit and interpretable fuzzy rule, following a Mamdani-style reasoning paradigm~\cite{815554,petrovic2006fuzzy,yang2017fuzzy}}, which consists of three fundamental steps: fuzzification, rule evaluation, and defuzzification. In this formulation, each fuzzy rule takes the canonical form:
\begin{equation}
\text{IF } x \text{ is } A_r \text{ THEN } y \text{ is } B_r,
\end{equation}
where \( x \in \mathbb{R}^D \) denotes the latent input, \( A_r \) is the fuzzy antecedent set, and \( B_r \) is the consequent. The antecedent is modeled using a Gaussian membership function centered at prototype \( c_r \) with width \( w_r \):
\begin{equation}
\label{eq:membership_function}
\mu_r(x) = \exp\left( - \frac{\|x - c_r\|^2}{2 w_r^2} \right).
\end{equation}

Instead of aggregating the outputs from all rules, we employ an OR-operation defuzzification strategy, wherein the rule with the highest membership value \( \mu^d_r(x) \) at dimension \(d\), determines the final modulation response as gating attention vector:

\begin{equation}
\label{eq:alpha}
\alpha^d = \max_r \mu^d_r(x).
\end{equation}

This approach enhances interpretability by providing a single, dominant decision pathway for each input, aligning with the core principles of Mamdani inference while maintaining computational tractability within a neural framework.

\subsection{iFuzzy Network Module}

To enable interpretable and flexible fuzzy reasoning, we define each fuzzy rule in terms of learnable prototype parameters in the raw input space. Specifically, each rule \( r \) is associated with a center prototype \( \tilde{c}_r \in \mathbb{R}^{D_{\text{raw}}} \) and a width prototype \( \tilde{w}_r \in \mathbb{R}^{D_{\text{raw}}} \). These prototypes are not directly used in inference; instead, they serve as semantically meaningful anchors in the raw EEG domain, allowing us to encode inductive biases (e.g., spectral priors) and enhance interpretability. Prior to fuzzy inference, these raw prototypes are transformed into the model’s latent space via a shared projector, yielding task-adaptive fuzzy representations. The detailed fuzzy gating pipeline is illustrated in Fig.~\ref{fig:fuzzy_detail}.

\revised{Unlike attention-based formulations, where similarity scores are computed between latent key and query vectors with no explicit physical meaning, the fuzzy membership in iFuzz-Meta is evaluated between projected EEG signals and rule prototypes that are explicitly defined in the raw EEG signal space. This design ensures that each rule activation corresponds to a concrete and interpretable neurophysiological pattern, rather than a latent similarity weight.}

\subsubsection{Fuzzy Rule Representation}

\revised{
Each fuzzy rule \( r \) is \emph{semantically defined in the original raw signal space} and \emph{operationally instantiated in the latent space} by projecting the raw prototypes through the projector \( p_0 \):
}
\begin{equation}
z = p_0(x), \quad c_r = p_0(\tilde{c}_r), \quad w_r = p_0(\tilde{w}_r).
\end{equation}
Here, \( z \in \mathbb{R}^{D} \) is the encoded input, and \( c_r, w_r \in \mathbb{R}^{D} \) represent the rule-specific fuzzy center and spread, respectively.
\revised{
Importantly, the latent fuzzy centers \( c_r \) and widths \( w_r \) are not free parameters learned independently in the embedding space, but deterministic projections of raw-space prototypes that correspond to interpretable neurophysiological patterns.
}
These transformed parameters \( (c_r, w_r) \) are the actual quantities used in subsequent fuzzy inference computations. This projection \( p_0 \) ensures compatibility between the input representation and rule parameters, while preserving a direct interpretive link to neuro-physiologically grounded features in the raw signal space.

\revised{
A shared projection function \(p_0\) is used for both inputs and fuzzy prototypes to maintain a consistent latent geometry for membership computation while preserving a traceable link between latent rule centers and their raw EEG definitions. Ablation details comparing with dual projection set are provided in Supplementary Section~II-E.}

\subsubsection{Rule Aggregation and Firing Computation}

For each input instance, we compute the membership of the encoded feature \( z = p_0(x) \in \mathbb{R}^{D} \) with respect to each fuzzy rule \( r \) using a Gaussian membership function, following eq. (\ref{eq:membership_function}):
\begin{equation}
\mu^d_{r}(z^d) = \exp\left( - \frac{\| z^d - c^d_r \|^2}{2 (w^d_r)^2 + \varepsilon} \right),
\end{equation}
where \( c_r = p_0(\tilde{c}_r) \) and \( w_r = p_0(\tilde{w}_r) \) denote the center and width of the \( r \)-th rule in the latent space, and \( \varepsilon \ll 1 \) is a small constant added for numerical stability. \(*^d\) denote any variable \(*\) at dimension \(d\). 

Unlike traditional Mamdani systems that aggregate outputs from all fuzzy rules into a composite inference surface, we adopt a streamlined OR-operation strategy to achieve sparse, interpretable reasoning. In this formulation, the firing strength of each rule is directly determined by its Gaussian membership value \(\mu^d_{r}(z^d)\). Following the eq. \ref{eq:alpha}, the rule at dimension \(d\) with the maximum activation is defined as:
\begin{equation}
\hat{r} = \arg\max_r \mu^d_{r}(z^d), \quad \alpha^d = \mu^d_{\hat{r}}(z^d).
\end{equation}

This approach implements a Mamdani-style \emph{max-membership defuzzification}, wherein the final decision is governed solely by the most confident rule. By restricting inference to a single activated prototype, the model avoids the semantic ambiguity that arises from blending multiple fuzzy outcomes. This not only simplifies interpretation, allowing clear attribution of decisions to specific rules, but also enhances computational efficiency during both training and inference.

Beyond interpretability, the OR-operation mechanism introduces desirable sparsity into the gradient flow, effectively reducing inter-rule redundancy and focusing learning dynamics on the most discriminative prototypes. As a result, the fuzzy module exhibits sharper attention over relevant regions of the latent space, promoting scalable and semantically grounded reasoning when integrated into deep neural architectures.

\subsubsection{Knowledge-Guided Modulation via Rule Activation}

Once the most relevant fuzzy rule \( r^\ast \) is selected, its firing strength \( \alpha = \mu_{r^\ast}(z) \) is used to modulate the input features through a soft gating mechanism. Specifically, we apply an element-wise multiplicative rescaling to the input tensor:
\begin{equation}
\label{eq:modulation}
z_{\text{output}} = z \odot \left(1 + \lambda \cdot \alpha \right),
\end{equation}
where \revised{\( \lambda =0.1\)} is a predefined modulation coefficient and \( \odot \) denotes element-wise multiplication. The scalar \( \alpha \in [0, 1] \) serves as a continuous gating signal derived from the confidence level of the fired fuzzy rule.

This formulation implements a form of fuzzy attention, where the modulation intensity is adaptively determined by the alignment between the input and the most activated fuzzy prototype. When \( \alpha \) is small, the modulated signal remains close to the original input, preserving the baseline representation. Conversely, a high activation \( \alpha \) amplifies features along dimensions strongly supported by the selected fuzzy rule, thereby emphasizing semantically relevant patterns in the neural signal.

Unlike traditional attention mechanisms, the proposed fuzzy-based modulation offers inherent interpretability by directly linking feature enhancement to explicit rule activations, enabling localized, semantically grounded modulation of neural representations.

Supplemental Material Section~I-B describes how the gradient of the iFuzzy Network Module is computed and applied during training.

\subsection{Loss Function Design}

The primary objective of our model is supervised classification, optimized via a standard cross-entropy (CE) loss computed between predicted logits and ground truth labels. The CE loss is defined as:
\begin{equation}
\mathcal{L}_{\mathrm{CE}} = - \sum_{i} y_i \log \hat{p}_i
\end{equation}
where \( y_i \) is the ground truth label and \( \hat{p}_i \) is the predicted probability for class \( i \). This formulation is used consistently across both meta-learning with two labels and fine-tuning stages with four labels.

To encourage desirable structure in the learned fuzzy parameters, we incorporate two auxiliary losses grounded in domain-informed inductive biases.

\subsubsection{Frequency-Regularized Center Loss}

To impose physiologically meaningful temporal structure on the learned fuzzy centers, we introduce a spectral regularization term that enforces a bandpass-like constraint in the frequency domain. Specifically, this loss penalizes both low- and high-frequency components, encouraging the fuzzy prototypes to concentrate their energy within a target frequency band typically associated with informative EEG rhythms (e.g., alpha and beta bands).

\begin{figure}[t]
    \centering
    \includegraphics[width=1\linewidth]{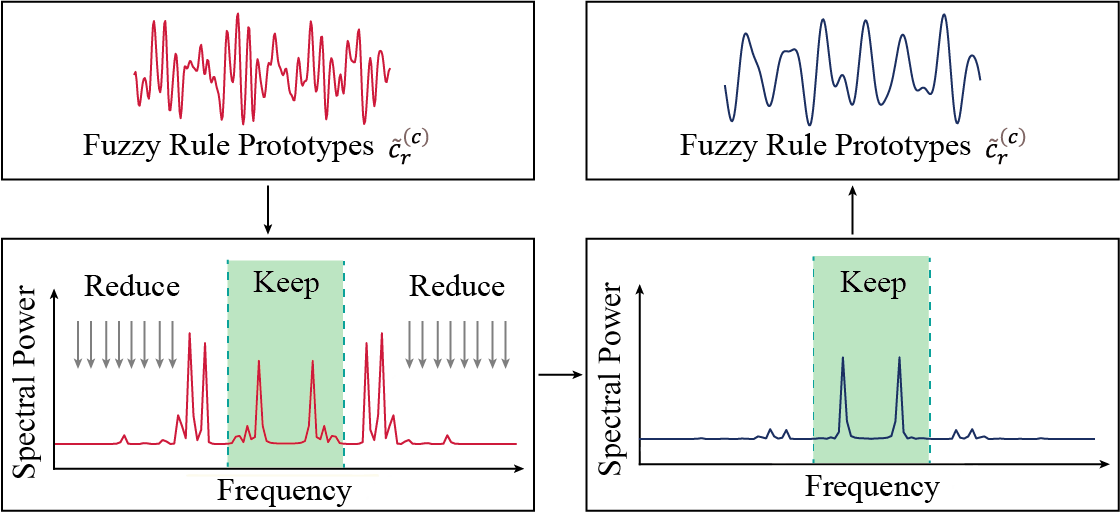}

    \caption{\revised{Illustration of the Frequency-Regularized Center Loss.
        The spectral energy of the fuzzy center is softly penalized outside the
        designated passband \( [f_{\min}, f_{\max}] \).
        This figure provides a conceptual illustration rather than results from
        real EEG data.
    }}

    \label{fig:fft_loss}
\end{figure}

\revised{
Let $\tilde{c}_r \in \mathbb{R}^{C \times T}$ denote the raw space fuzzy center of rule $r$, where $C$ is the number of channels and $T$ is the number of time points. We compute its discrete Fourier transform, denoted as $\hat{c}_r(f)$, and define the frequency regularization loss as
\begin{equation}
\label{eq:loss_freq}
\mathcal{L}_{\mathrm{freq}} = \frac{1}{R} \sum_{r=1}^R \sum_{f \notin [f_{\min}, f_{\max}]} \omega(f)\, \bigl|\hat{c}_r(f)\bigr|^2 ,
\end{equation}
where $[f_{\min}, f_{\max}]$ specifies the target passband and $\omega(f)$ is a frequency dependent weighting function. Fig.~\ref{fig:fft_loss} provides a conceptual illustration of the spectral penalty imposed outside the passband.

The weighting function $\omega(f)$ is designed as a symmetric, band aware penalty. Frequencies within the target band are not penalized, while frequencies outside the band are softly penalized according to their distance from the nearest band boundary. Specifically,
\begin{equation}
\omega(f)=
\begin{cases}
\dfrac{f_{\min}-f}{f_{\min}-f_{\mathrm{low}}}, & f_{\mathrm{low}} \le f < f_{\min}, \\[6pt]
0, & f_{\min} \le f \le f_{\max}, \\[6pt]
\dfrac{f-f_{\max}}{f_{\mathrm{Nyq}}-f_{\max}}, & f_{\max} < f \le f_{\mathrm{Nyq}},
\end{cases}
\end{equation}
where $f_{\mathrm{low}}$ denotes the minimum analyzed frequency, fixed at $0.5$~Hz in this study, and $f_{\mathrm{Nyq}}$ is the Nyquist frequency.
}

This spectral loss explicitly incorporates neuroscientific priors by shaping each fuzzy center to suppress noise-dominated frequencies outside the functional EEG range while preserving components known to reflect cognitive activity. In contrast to conventional architectures that implicitly learn frequency characteristics through convolutional filters, our formulation offers a transparent and tunable pathway for embedding domain knowledge into the learning process. As we demonstrate in later analysis, this constraint leads to the emergence of physiologically interpretable patterns, especially in the frequency and spatial domains, that align with known EEG characteristics. Supplemental Material Section~I-C describes how the gradient of this loss is computed and applied during training.

\paragraph{Inductive Bias and Domain-Aware Knowledge Transfer}

The frequency-regularized loss serves as a principled mechanism to inject domain knowledge into the model by explicitly constraining the spectral profile of each fuzzy center. Rather than relying solely on data-driven optimization, we enforce an inductive bias that aligns learned prototypes with established EEG characteristics. Specifically, we apply a frequency-domain penalty that suppresses spectral components outside the 0.5-8 Hz range, which is commonly associated with slow cortical potentials and intention-related neural dynamics in visual cognitive tasks, specially delta activity in distinction between OR and OI\cite{10570151}, and slow cortical potentials (SCPs)~\cite{Koenig2025Spontaneous}.

This biologically grounded constraint enables the model to prioritize neural patterns that are known to be stable across individuals, thereby improving generalization under cross-subject and few-shot conditions. By guiding fuzzy rule formation toward shared low-frequency structures, the model avoids overfitting to subject-specific noise and instead extracts semantically meaningful, task-relevant features. As a result, the fuzzy module becomes not only interpretable but also robust and transferable, key properties for real-world BCI systems operating under non-stationary and low-data regimes.

\subsubsection{Total loss}  
The final loss is a weighted sum of all components:

\begin{equation}
\mathcal{L}_{\text{total}} = \mathcal{L}_{\text{CE}} + \lambda_{\text{freq}} \cdot \mathcal{L}_{\text{freq}} + \lambda_{\text{div}} \cdot \mathcal{L}_{\text{div}},
\end{equation}

where \( \lambda_{\text{freq}} \) and \( \lambda_{\text{div}} \) (diversity loss, details are shown in Supplemental Material Section~I-D) control the strength of the frequency regularization and diversity prior, defined as 0.3, respectively. These auxiliary terms introduce weak structural guidance while preserving the flexibility of the fuzzy representation.

\revised{Although the total objective is expressed as a weighted sum, the proposed integration is realized through coupled gradients acting on the same prototype parameters.
The frequency regularization introduces prior-informed gradients, while task supervision provides empirical gradients, allowing data-driven learning to modulate the effective influence of the prior during optimization.}

\section{Results}
\label{sec:results}

We evaluate the classification performance of the proposed iFuzz-Meta under varying few-shot ratios (FSR = 0.1 and 0.3), model architectures (TSPA, TSpPA, TSSPA), and training regimes (with and without meta-learning, denoted as \textit{wMeta}). All experiments are conducted using the same experimental setup as EEGNet\revised{~\cite{lawhernEEGNetCompactConvolutional2018}}, which serves as \revised{one of} the baseline model. \revised{Other deep learning baseline models are EEGformer~\cite{eegformer} and EEG-TCNet~\cite{9283028}, as well as two Deep FNN models~\cite{jiang2024ifuzzytl,10849810}.} Statistical significance is assessed using paired t-tests\revised{, with multiple-comparison correction performed using the Benjamini--Hochberg false discovery rate (FDR-BH) procedure}. 

\subsection{Meta-Learning Performance Evaluation}

We first assess the effectiveness of the meta-learning phase under two few-shot settings (FSR = 0.1 and 0.3). As shown in Table~\ref{tab:meta_learning}, all three iFuzz-Meta structures (TSPA, TSpPA, TSSPA) significantly outperform EEGNET and EEGformer ($p < 0.05$) during meta-learning across both FSR levels, but no significant difference with EEG-TCNet ($p > 0.05$). There is no significant differences observed among the iFuzz-Meta variants themselves (all $p > 0.05$, paired t-tests), nor between FSR = 0.1 and FSR = 0.3 for the same architecture (all $p > 0.05$). These results suggest that increasing the number of support samples has limited impact on the effectiveness of the meta-learning phase.

\begin{table*}[t!]
\centering
\scriptsize
\caption{
Meta-learning accuracy (\%) [2 categories, OI vs OR] of EEGformer, EEG-TCNet, EEGNet, and iFuzz-Meta under different FSRs (Number of Rules=4).
}
\label{tab:meta_learning}
\begin{tabular}{c c c c c c c}
\toprule
FSR & EEGformer & EEG-TCNet & EEGNet & \multicolumn{3}{c}{iFuzz-Meta} \\
\cmidrule(lr){5-7}
 &  &  &  & TSPA & TSpPA & TSSPA \\
\midrule
0.1 & 74.97 ± 21.13$^{*}$ & 81.49 ± 22.97 & 73.96 ± 20.84$^{*}$ & 85.30 ± 14.06 & 87.84 ± 12.82 & 87.92 ± 12.65 \\
0.3 & 74.80 ± 21.10$^{*}$ & 82.60 ± 23.30 & 73.88 ± 20.84$^{*}$ & 85.11 ± 14.18 & 87.93 ± 12.37 & 87.53 ± 12.88 \\
\bottomrule
\end{tabular}
\begin{minipage}{0.9\linewidth}
\footnotesize
\textbf{Note.} Values are mean ± standard deviation. 
Significance markers ($^*$) indicate that baselines (EEGformer, TCNet, EEGNet) are significantly worse than iFuzz-Meta ($p<0.05$).  
No significant differences were found among iFuzz-Meta structures (TSPA, TSpPA, TSSPA; $p>0.05$).
\end{minipage}
\end{table*}

\subsection{Overall Performance Comparison}

We further compare the end-to-end classification performance of iFuzz-Meta and three baseline models, EEGformer, EEG-TCNet, and EEGNet, across the two paradigms (OI and OR), under both meta-learning and non-meta-learning conditions. Results are summarized in Table~\ref{tab:fsr01_nr4} (FSR = 0.1) and Table~\ref{tab:fsr03_nr4} (FSR = 0.3), with the number of fuzzy rules fixed at 4.

In both FSR settings, iFuzz-Meta consistently outperforms EEGformer, EEG-TCNet, and EEGNet under with meta-learning conditions (all $p < 0.05$), demonstrating the benefit of incorporating interpretable fuzzy rules and cross-task knowledge adaptation. In contrast, without meta-learning (i.e., models trained directly on support sets), the performance of iFuzz-Meta is higher than EEG-TCNET (all $p < 0.05$) except TSSPA in OR task with FSR = 0.1($p > 0.05$). A significant improvement is observed in the OR task under FSR = 0.1, where the TSPA framework of iFuzz-Meta achieves $77.76\% \pm 6.47\%$, significantly higher than EEGNet’s $71.78\% \pm 8.33\%$ ($p < 0.001$).

When comparing training modes (with vs. without meta-learning), we observe that meta-learning generally leads to improved performance. Specifically, under FSR = 0.1, most model-task combinations benefit from meta-learning, specially TSpPA and TSSPA in OR task, where show significant differences($p < 0.05$). Under FSR = 0.3, the improvements are more selective; only TSpPA show no significant gains in the OI task ($p > 0.05$). These findings suggest that while increasing the support set size contributes to better fine-tuning, meta-learning remains a critical component in achieving consistent and robust classification performance.

\revised{Comparisons with deep fuzzy neural networks (FNNs) are provided in Supplemental Material Section~II-A, while the convergence analysis is presented in Supplemental Material Section~II-C, demonstrating that the proposed framework achieves both interpretability and competitive performance.}

\begin{table*}[htbp]
\centering
\scriptsize
\caption{
Overall Accuracy (\%) [4 categories] of EEGformer, EEG-TCNet, EEGNet, and iFuzz-Meta (FSR = 0.1, Number of Rules = 4) across different tasks, model layers, and meta-learning conditions.
}

\label{tab:fsr01_nr4}
\begin{tabular}{cclllll}
\toprule
\textbf{Task} & \textbf{\revised{Model Arch}} & \textbf{wMeta} & \textbf{EEGformer} & \textbf{EEG-TCNet} & \textbf{EEGNet} & \textbf{iFuzz-Meta} \\
\midrule
\multirow{6}{*}{OI} 
    & \multirow{2}{*}{TSPA}  & False & 69.49 ± 5.24 & 69.56 ± 7.16$^{*}$ & 70.73 ± 7.28 & \textbf{74.09 ± 8.27} \\
    &                        & True  & 70.38 ± 6.62$^{***}$ & 69.89 ± 6.57$^{***}$ & 71.35 ± 6.71$^{**}$ & \textbf{76.82 ± 4.60} \\
    & \multirow{2}{*}{TSpPA} & False & 69.49 ± 5.24 & 69.56 ± 7.16$^{*}$  & 70.73 ± 7.28 & \textbf{73.89 ± 7.60} \\
    &                        & True  & 70.38 ± 6.62$^{***}$ & 69.89 ± 6.57$^{***}$ & 71.35 ± 6.71$^{**}$ & \textbf{76.98 ± 4.71} \\
    & \multirow{2}{*}{TSSPA} & False & 69.49 ± 5.24$^{*}$ & 69.56 ± 7.16$^{*}$  & 70.73 ± 7.28 & \textbf{73.68 ± 8.29} \\
    &                        & True  & 70.38 ± 6.62$^{***}$ & 69.89 ± 6.57$^{***}$ & 71.35 ± 6.71$^{**}$ & \textbf{77.14 ± 4.97} \\
\midrule
\multirow{6}{*}{OR} 
    & \multirow{2}{*}{TSPA}  & False & 68.09 ± 6.89$^{***}$ & 67.34 ± 6.81$^{***}$ & 68.87 ± 6.96$^{***}$ & \textbf{77.76 ± 6.47} \\
    &                        & True  & 71.46 ± 5.63$^{***}$ & 71.17 ± 5.61$^{***}$ & 72.14 ± 5.69$^{***}$ & \textbf{77.87 ± 4.74} \\
    & \multirow{2}{*}{TSpPA} & False$^{*}$ & 68.09 ± 6.89$^{*}$ & 67.34 ± 6.81$^{*}$ & 68.87 ± 6.96 & \textbf{73.26 ± 6.95} \\
    &                        & True  & 71.46 ± 5.63$^{***}$ & 71.17 ± 5.61$^{***}$ & 72.14 ± 5.69$^{**}$ & \textbf{76.77 ± 4.32} \\
    & \multirow{2}{*}{TSSPA} & False$^{*}$ & 68.09 ± 6.89 & 67.34 ± 6.81 & 68.87 ± 6.96 & \textbf{73.61 ± 7.10} \\
    &                        & True  & 71.46 ± 5.63$^{***}$ & 71.17 ± 5.61$^{***}$ & 72.14 ± 5.69$^{**}$ & \textbf{76.99 ± 4.35} \\
\bottomrule
\end{tabular}

\vspace{0.5em}
\begin{minipage}{0.92\linewidth}
\footnotesize
\textbf{Note.} Values are mean ± standard deviation. Boldface indicates the highest accuracy in each row. 
Significance markers $^*$, $^{**}$, and $^{***}$ indicate $p < 0.05$, $p < 0.01$, and $p < 0.001$, respectively, reflecting differences compared to iFuzz-Meta. 
wMeta significance markers indicate iFuzz-Meta comparisons with Meta-learning vs without Meta-learning. FSR denotes the few-shot ratio. \revised{Arch denotes Architecture.}
\end{minipage}
\end{table*}

\begin{table*}[htbp]
\centering
\scriptsize
\caption{
Overall Accuracy (\%) [4 categories] of EEGformer, EEG-TCNet, EEGNet, and iFuzz-Meta (FSR = 0.3, Number of Rules = 4) across different tasks, model layers, and meta-learning conditions.
}

\label{tab:fsr03_nr4}
\begin{tabular}{cclllll}
\toprule
\textbf{Task} & \textbf{\revised{Model Arch}} & \textbf{wMeta} & \textbf{EEGformer} & \textbf{EEG-TCNet} & \textbf{EEGNet} & \textbf{iFuzz-Meta} \\
\midrule
\multirow{6}{*}{OI} 
    & \multirow{2}{*}{TSPA}  & False$^{*}$ & 92.12 ± 7.45$^{*}$ & 92.09 ± 7.44$^{*}$ & 93.22 ± 7.54 & \textbf{95.72 ± 4.62} \\
    &                        & True  & 93.36 ± 5.73$^{***}$ & 93.01 ± 5.71$^{***}$ & 93.96 ± 5.76$^{**}$ & \textbf{97.90 ± 2.26} \\
    & \multirow{2}{*}{TSpPA} & False & 92.12 ± 7.45 & 92.09 ± 7.44 & 93.22 ± 7.54 & \textbf{94.58 ± 6.82} \\
    &                        & True  & 93.36 ± 5.73$^{**}$ & 93.01 ± 5.71$^{***}$ & 93.96 ± 5.76$^{*}$ & \textbf{97.04 ± 2.53} \\
    & \multirow{2}{*}{TSSPA} & False$^{*}$ & 92.12 ± 7.45 & 92.09 ± 7.44$^{*}$ & 93.22 ± 7.54 & \textbf{95.40 ± 4.88} \\
    &                        & True  & 93.36 ± 5.73$^{**}$ & 93.01 ± 5.71$^{***}$ & 93.96 ± 5.76$^{**}$ & \textbf{97.68 ± 2.34} \\
\midrule
\multirow{6}{*}{OR} 
    & \multirow{2}{*}{TSPA}  & False$^{**}$ & 92.15 ± 7.24 & 91.64 ± 7.20$^{*}$ & 93.24 ± 7.33 & \textbf{95.45 ± 4.61} \\
    &                        & True  & 93.12 ± 5.90$^{***}$ & 92.10 ± 5.84$^{***}$ & 94.12 ± 5.96$^{**}$ & \textbf{98.28 ± 1.96} \\
    & \multirow{2}{*}{TSpPA} & False$^{*}$ & 92.15 ± 7.24$^{*}$ & 91.64 ± 7.20$^{**}$ & 93.24 ± 7.33 & \textbf{94.77 ± 5.64} \\
    &                        & True  & 93.12 ± 5.90$^{***}$ & 92.10 ± 5.84$^{***}$ & 94.12 ± 5.96$^{**}$ & \textbf{98.04 ± 2.56} \\
    & \multirow{2}{*}{TSSPA} & False$^{*}$ & 92.15 ± 7.24$^{*}$ & 91.64 ± 7.20$^{**}$ & 93.24 ± 7.33 & \textbf{94.81 ± 6.48} \\
    &                        & True  & 93.12 ± 5.90$^{**}$ & 92.10 ± 5.84$^{***}$ & 94.12 ± 5.96$^{**}$ & \textbf{97.89 ± 2.25} \\
\bottomrule
\end{tabular}

\vspace{0.5em}
\begin{minipage}{0.92\linewidth}
\footnotesize
\textbf{Note.} Values are mean ± standard deviation. Boldface indicates the highest accuracy in each row. 
Significance markers $^*$, $^{**}$, and $^{***}$ indicate $p < 0.05$, $p < 0.01$, and $p < 0.001$, respectively, reflecting differences compared to iFuzz-Meta. 
wMeta significance markers indicate iFuzz-Meta comparisons with Meta-learning vs without Meta-learning. FSR denotes the few-shot ratio. \revised{Arch denotes Architecture.}
\end{minipage}
\end{table*}

\subsection{Ablation Study}
We conduct two ablation studies: one assessing the contribution of the frequency-regularized center loss, and other examining the impact of the number of rules (Supplemental Material Section~II-\revised{B}1) \revised{and comparing with the centroid-based soft defuzzification (Supplemental Material Section~II-\revised{B}2)}. 

To evaluate the effectiveness of the frequency-regularized center loss, we analyze how incorporating neuroscientific priors influences the behavior of fuzzy prototypes by constraining them in the frequency domain. As shown in Table~\ref{tab:freq_loss}, this regularization consistently improves accuracy under FSR = 0.3 across all tasks and model structures ($p<0.001$). 

Under the more challenging FSR = 0.1 setting, where the training data is scarce, improvements are observed in some cases (e.g., TSSPA structure in OI Task: +1.41\%, $p<0.01$), while others remain comparable or slightly lower ($p>0.05$). This suggests that the benefit of frequency-domain priors becomes more prominent as the support set grows.

Overall, these results validate the proposed loss as an effective mechanism for embedding domain knowledge into the fuzzy representation. Unlike conventional architectures that rely purely on data-driven learning, our approach enables interpretable and physiologically meaningful regularization, improving generalization especially when sufficient data is available.

\begin{table}[ht]
\centering
\scriptsize
\caption{
Accuracy (\%) of iFuzzy with and without knowledge-based frequency-regularized center loss, across different few-shot ratios (FSR), task types, and model layers.
}
\label{tab:freq_loss}
\begin{tabular}{cccll}
\toprule
\textbf{FSR} & \textbf{Task Type} & \textbf{\revised{Model Arch}} & \textbf{No Frequency} & \textbf{Frequency} \\
\midrule
\multirow{6}{*}{0.1} 
 & \multirow{3}{*}{OI} 
    & TSPA   & \textbf{76.96 ± 0.95} & 76.82 ± 0.92 \\
 &  & TSpPA  & 75.89 ± 1.10 & \textbf{76.98 ± 0.94} \\
 &  & TSSPA  & 75.73 ± 1.10 & \textbf{77.14 ± 0.99}$^{**}$ \\
\cmidrule{2-5}
 & \multirow{3}{*}{OR} 
    & TSPA   & 76.94 ± 0.85 & \textbf{77.87 ± 0.95} \\
 &  & TSpPA  & \textbf{76.99 ± 0.84} & 76.77 ± 0.86 \\
 &  & TSSPA  & \textbf{77.14 ± 0.91} & 76.99 ± 0.87 \\
\midrule
\multirow{6}{*}{0.3} 
 & \multirow{3}{*}{OI} 
    & TSPA   & 95.37 ± 0.28 & \textbf{97.90 ± 0.45}$^{***}$ \\
 &  & TSpPA  & 95.23 ± 0.27 & \textbf{97.04 ± 0.51}$^{***}$ \\
 &  & TSSPA  & 95.25 ± 0.31 & \textbf{97.68 ± 0.47}$^{***}$ \\
\cmidrule{2-5}
 & \multirow{3}{*}{OR} 
    & TSPA   & 95.24 ± 0.31 & \textbf{98.28 ± 0.39}$^{***}$ \\
 &  & TSpPA  & 94.99 ± 0.32 & \textbf{98.04 ± 0.51}$^{***}$ \\
 &  & TSSPA  & 95.23 ± 0.34 & \textbf{97.89 ± 0.45}$^{***}$ \\
\bottomrule
\end{tabular}

\vspace{0.5em}
\begin{minipage}{0.9\linewidth}
\footnotesize
\textbf{Note.} Values are mean ± standard deviation. Boldface indicates the higher accuracy between the two settings in each row. 
$^*$, $^{**}$, and $^{***}$ indicate significant improvement over No-Frequency loss condition from with-Frequency loss condition ($p<0.05$, $p<0.01$, $p<0.001$). \revised{Arch denotes Architecture.}
No significant differences (all $p>0.05$) were found among iFuzz-Meta structures (TSPA, TSpPA, and TSSPA). 
\end{minipage}
\end{table}

\subsection{Explanation and Interpretability}

\subsubsection{Interpretable Representations through Architecture Design}

To better understand how architectural design shapes the representations learned by iFuzz-Meta, we examine the spectral power (0.5-8Hz) of fuzzy center prototypes from three frameworks: TSPA, TSpPA, and TSSPA, as described in Section \ref{sec:model_arch}. Each framework encodes prior knowledge via different combinations of temporal and spatial pattern analysis.

As shown in Fig. \ref{fig:model_structures}(D), TSPA, which emphasizes temporal convolution early in the pipeline, tends to learn center representations dominated by temporal structure with limited spatial differentiation. TSpPA, by incorporating spatial aggregation earlier, captures smoother and more coherent spatial activations. TSSPA, which sequentially combines spatial and temporal processing, learns center patterns that exhibit both spatial precision and temporal dynamics. These representations often resemble canonical EEG structures such as ERPs, suggesting a closer alignment with neurophysiological patterns.

While all three architectures achieve similar classification performance, TSSPA yields more structured and interpretable prototypes. This suggests that well-designed architectural priors can support more meaningful rule learning. Such interpretability not only aids in understanding the internal representations of the model but also bridges the gap between data-driven deep learning and neuroscientific theory, offering pathways for explanation in neural decoding tasks.

\subsubsection{Center Shifting from Meta-Learning to Fine-Tuning}

The fuzzy center prototypes learned during meta-training exhibit more dispersed and irregular spectral-spatial patterns, as seen in the top row of Fig.~\ref{fig:center_shifting}(A). This reflects the task-agnostic nature of meta-learning, where the objective is to capture transferable priors rather than task-specific features. Such variability is expected, as the model is exposed to both OI and OR meta-tasks, which differ in granularity and abstraction.

After subject-specific fine-tuning, these centers converge into smoother and more spatially consistent patterns, resembling the post-hoc interpretability results reported in~\cite{leong2023ventral}. This suggests that fine-tuning successfully adapts the generic meta-learned prior into task-specific spatial filters. From a neuroscience perspective, the emergence of localized and smoother activity aligns with the known spatial organization of cognitive tasks in the brain \cite{nunez2006electric}. Notably, the smoothness is more prominent in OI compared to OR, indicating that fine-grained identification may rely more on spatial encoding, whereas OR benefits from broader semantic abstraction.

These findings support the effectiveness of TSSPA in enabling a two-phase learning process: general representation acquisition followed by task specialization.

\subsubsection{Prototype Differences With vs. Without Meta-Learning}

\begin{figure*}[htbp]
    \centering
    \includegraphics[width=1\linewidth]{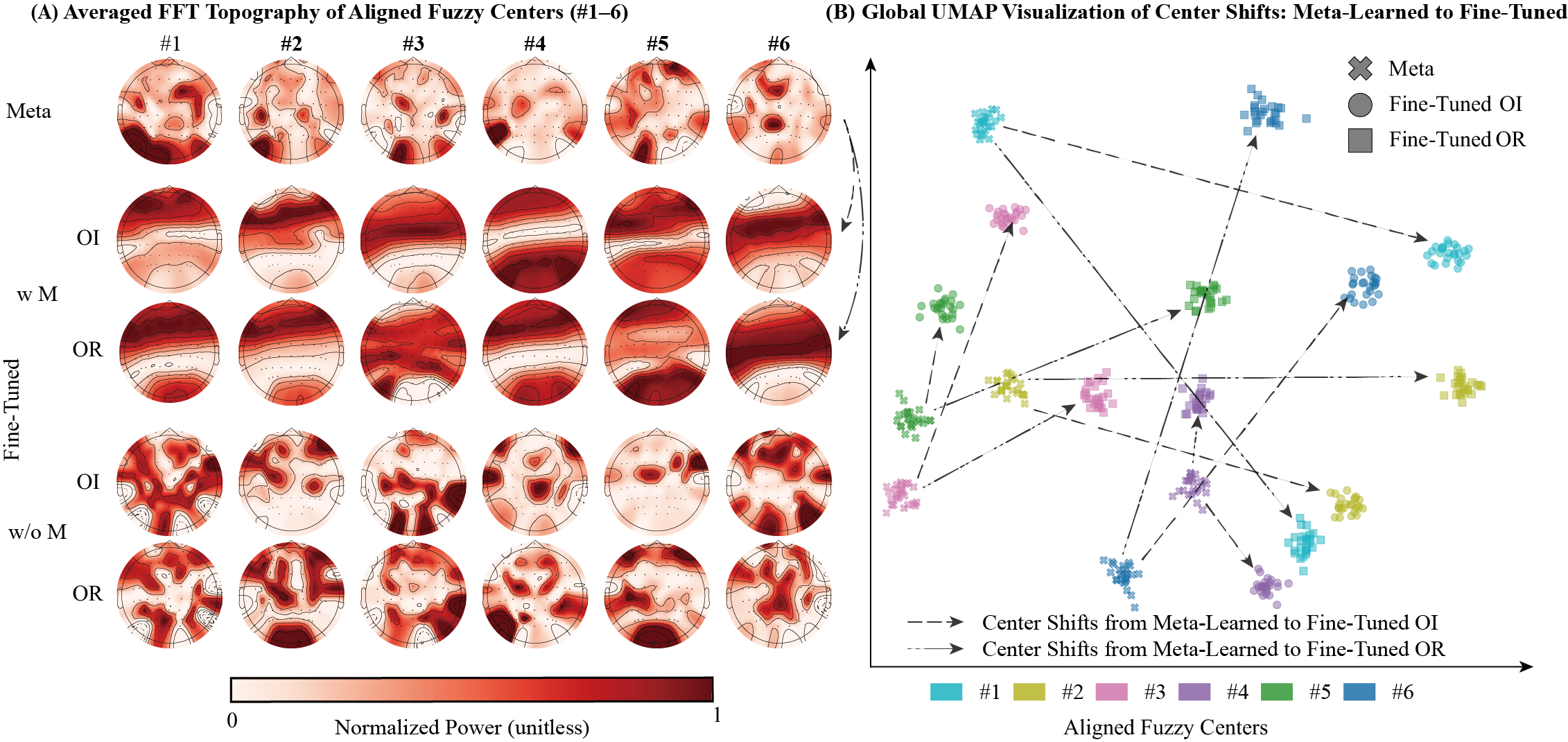}

    \caption{
    \textbf{(A) Averaged topographic maps} of the aligned fuzzy center prototypes (0.5-8 Hz power from the TSSPA model using six fuzzy rules.) Rows represent different learning stages and conditions: Meta-learning, Fine-Tuning with Meta (w M), and Fine-Tuning without Meta (w/o M), each for Object Identification (OI) and Object Recognition (OR). Meta-learned centers serve as initialization and evolve into smoother, more structured representations through fine-tuning.
    \textbf{(B) Global UMAP visualization of fuzzy center shifts from meta-learning to fine-tuned models.} Each dot represents a fuzzy rule from a model evaluated on a specific subject. Color denotes rule OI (\#1-6), while marker shape indicates the model stage. Cosine similarity is used for aligning centers. Solid arrows show shifts toward Fine-Tuned OI; dashed arrows indicate shifts toward Fine-Tuned OR.
    }
    \label{fig:center_shifting}
\end{figure*}

To assess the impact of meta-learning, we compare the fuzzy center topographies from models with and without the meta-learning phase. Without meta-learning, fine-tuning directly from scratch results in spatially fragmented and inconsistent center patterns, as shown in the bottom row of Fig.~\ref{fig:center_shifting}(A). This suggests that the model struggles to organize the latent space effectively when deprived of a structured initialization.

In contrast, when initialized via meta-learning, the subsequent fine-tuning produces more coherent and interpretable prototypes. This confirms that meta-learned priors facilitate more stable optimization trajectories and encourage the reuse of cognitively meaningful patterns across tasks.

The difference is further evident in the UMAP visualization in Fig.~\ref{fig:center_shifting}(B), where the meta-learned centers serve as attractors for fine-tuned representations. Without such anchors, the distribution becomes more scattered, indicating weaker alignment across subjects or tasks. These observations underscore the utility of meta-learning not only for generalization but also for promoting structural consistency in the learned representations.

\revised{To further strengthen the empirical support for the proposed framework, we conducted an additional out-of-domain validation experiment using a public, multi-day, high-quality motor imagery (MI) EEG dataset~\cite{Yang2025MIDataset}. This dataset represents a neural paradigm fundamentally different from the original OI/OR tasks. Detailed experimental settings and results are provided in Supplemental Material Section~II-D. The results show the consistent better performance of the iFuzz-Meta comparing to the baselines.}

\section{Discussion}
\label{sec:discussion}

\subsection{Neural Interpretation of Fuzzy Center Prototypes}

To investigate the neurophysiological knowledge of the learned fuzzy representations, we visualize the aligned fuzzy center prototypes from the TSSPA architecture in both OI and OR tasks (Fig.~\ref{fig:model_structures}(D)). Each fuzzy rule center corresponds to a spatiotemporal prototype capturing dominant EEG patterns, with the topographic map representing the spectral power (0.5--8~Hz) and the waveform illustrating temporal dynamics.

In the OI task, the learned centers exhibit spatial selectivity aligned with primary sensory processing regions. Notably, rule~\#3 shows dominant activation over occipital sites (Oz, POz, O1/2), consistent with early-stage visual processing in the primary visual cortex~\cite{diRusso2003visual}. Rule~\#2 displays parietal activity with mid-latency dynamics, potentially associated with spatial attention mechanisms~\cite{luck1997attention}. Rule~\#1 emphasizes central-parietal channels (Cz, CPz) with sustained temporal structure, which may reflect preparatory motor or attentional gating processes~\cite{hillyard1998gain}. In contrast, the OR task reveals a different organization of fuzzy rules. Rule~\#2 exhibits strong parietal activation with a \revised{P300-like temporal profile}, \revised{which is qualitatively consistent with neural processes related to semantic evaluation and decision formation}~\cite{polich2007p300}. Rules~\#1 and~\#3 also recruit parietal regions but exhibit distinct temporal profiles, suggesting complementary roles in feature accumulation and categorical abstraction.

\revised{Overall, these observations indicate that iFuzz-Meta learns neurophysiologically plausible spatiotemporal prototypes whose structure is consistent with established cognitive components.} The interpretability of the fuzzy rule centers, which manifested through spatially localized and temporally distinct prototypes, highlights the model’s capacity to uncover human-understandable features from EEG, bridging decoding accuracy with neurophysiological insight.

\subsection{Center Prototype Consistency and Microstate--Level Interpretation}

As shown in Fig.~\ref{fig:center_shifting}(B), Rule~\#4 exhibits the smallest adaptation distance across the meta-learning, OR, and OI tasks, suggesting that this fuzzy rule remains highly stable under domain adaptation. The corresponding topographic distribution in Fig.~\ref{fig:center_shifting}(A) displays a bilateral fronto-parietal activation band with a midline transition, characteristic of EEG microstate~C \cite{tarailis2024functional,asha2024analysis}. This microstate has been consistently linked to higher-order cognitive functions and self-referential processing.

From a neuro-scientific perspective, microstate~C has been associated with the integration of self-referential and personally significant information, potentially bridging internally oriented mentation and externally directed cognitive control~\cite{Tarailis2023}. It has also been shown to interact with both the default mode and cognitive control systems~\cite{Seitzman2017}, and to exhibit functional connectivity with the cingulo-opercular, dorsal attention (DAN), and default mode networks (DMN) at rest~\cite{Croce2018}. Collectively, these findings indicate that microstate~C may act as a transitional or integrative neural state supporting the coordination between semantic evaluation and goal-directed individuation. In the context of the present OR--OI paradigm, its persistence across tasks implies a stable control mechanism that maintains cognitive continuity as processing shifts from category-level recognition to instance-level specification.

From an algorithmic viewpoint, the minimal domain-shift of Rule~\#4 demonstrates that the iFuzz-Meta framework captures a task-invariant fuzzy prototype aligned with this stable neurocognitive microstate. The fuzzy rule center therefore functions as a shared anchor in the rule space, preserving interpretable reasoning while adapting to task variations. This alignment between algorithmic stability and neurophysiological consistency suggests that iFuzz-Meta not only achieves cross-subject robustness but also mirrors invariant control states of the human brain, providing a mechanistic bridge between interpretable fuzzy logic and dynamic neural organization.

\subsection{Hierarchical Task Design Enables Intention-Aware Meta-Learning}

Cognitive neuroscience has long emphasized that visual object processing unfolds hierarchically, progressing from coarse semantic categorization to fine-grained individuation of specific exemplars. In this framework, OR primarily engages semantic-level categorization supported by higher-order association areas, whereas OI demands individuating specific targets through occipito-parietal attentional mechanisms and ventral-stream perceptual detail. This transition from OR to OI has been interpreted as a coarse-to-fine processing strategy, whereby intentional, goal-directed attention progressively refines categorical representations into individuated ones~\cite{turn0search4}. 

Motivated by this cognitive hierarchy, we formulate the OR→OI transition as a binary meta-task. Instead of treating OR and OI as two independent four-way classification problems, we explicitly model the shift between semantic recognition and individuation as a higher-level task. This cognitively grounded design encourages the network to extract intention-relevant features that capture the representational refinement from category to instance. The resulting meta-learned features are then reused to support downstream four-way decoding in both OR and OI settings.

This hierarchical task structure provides multiple benefits. First, it improves sample efficiency by guiding the model to focus on discriminative features that generalize across subjects. Second, it enhances interpretability: during meta-training, fuzzy prototypes align with distinct spatial-temporal patterns consistent with semantic versus individuation stages, providing neurocognitively meaningful insight into intention-related processing. Finally, by bridging intention modeling with class-level decoding, the framework not only achieves superior performance but also offers a principled pathway toward intention-aware BCI applications.

\subsection{Balancing Interpretability and Model Capacity}

A central design goal of iFuzz-Meta is to enhance interpretability via rule-based modulation and prototype-based reasoning. Unlike conventional deep networks that rely on opaque high-dimensional feature transformations, our framework constrains decision-making through a limited set of fuzzy rules, each associated with human-interpretable spatiotemporal EEG prototypes. These rules serve as semantic anchors that not only drive classification but also offer neuro-scientific insights into the latent structure of neural representations.

This interpretability is especially valuable in BCI applications, where understanding 'How' a decision is made is often as important as the decision itself, particularly in clinical or cognitive neuroscience contexts. The fuzzy module in our architecture bridges this gap by providing a symbolic layer that translates distributed neural signals into rule-driven reasoning, yielding explanations that are both data-aligned and theory-consistent.

Importantly, our framework aligns with the emerging view that neuroscience is constrained by imperfect theories~\cite{varoquaux2021ai}, where fully predictive models of cognition are lacking due to the complexity of the brain and limitations in current measurement tools. In this setting, purely data-driven models may overfit to noise or spurious correlations, while purely theory-driven models often fail to generalize. iFuzz-Meta offers a principled compromise by integrating both \emph{bottom-up} (data-driven) learning and \emph{top-down} (theory-regularized) inductive biases.

Concretely, the fuzzy centers are initialized and regularized using physiologically motivated priors, such as frequency-domain constraints that favor meaningful EEG rhythms (e.g., alpha or beta bands), and diversity losses that encourage the separation of functional motifs. These priors do not impose rigid structure but guide the learning trajectory toward solutions that are more biologically plausible and transferable across subjects. At the same time, the centers and rules are refined through gradient-based optimization, allowing the model to discover emergent patterns not captured by existing theories.

Beyond interpretability, the low-dimensional and smoothly evolving rule activations produced by iFuzz-Meta are particularly well suited for real-time deployment. They can serve as stable control signals for online BCI operation, continuous cognitive-state monitoring, or adaptive neurofeedback, offering transparency and robustness that are difficult to achieve with purely deep architectures.

\revised{Although the Gaussian membership function used in iFuzz-Meta is mathematically similar to kernel-based similarity measures, the proposed framework is not an attention mechanism. In attention models, key and query vectors are optimized solely for discriminative similarity and do not correspond to any physically interpretable EEG structure. In contrast, iFuzz-Meta defines each fuzzy rule as a neurophysiologically grounded prototype in the raw signal space, and inference is performed as rule attribution rather than feature aggregation.}

\subsection{Limitations}
While iFuzz-Meta enhances clarity and prototype distinctiveness, it may underrepresent synergistic relationships among rules. In complex or overlapping domains, such sparsity could constrain representational flexibility by neglecting secondary yet informative contributions. Future extensions may incorporate hierarchical or multi-rule routing to balance interpretability with expressive power.

\revised{
\subsection{Future Work}

Several directions remain open for future investigation.
First, the proposed iFuzz-Meta framework can be extended beyond EEG to other neuroimaging modalities, such as functional near-infrared spectroscopy and functional magnetic resonance imaging, to assess its generality across measurement principles and spatiotemporal resolutions.
Second, while the current design adopts single-rule activation for interpretability, future work may explore multi-rule synergy mechanisms that preserve transparent and human-understandable reasoning.
Third, richer forms of knowledge integration may be investigated by incorporating domain priors beyond frequency information, including spatial connectivity and temporal dynamics.
Finally, future work will explore statistically grounded validation strategies for prototype–neurophysiology alignment, as well as system-level optimization, especially zero-shot transfer learning, for fully online BCI deployment, including low-latency inference and streaming adaptation. 
}

\section{Conclusion}
\label{sec:conclusion}

This study introduced iFuzz-Meta, a general interpretable fuzzy learning framework that unifies rule-based reasoning with adaptive meta-learning. The framework preserves human-understandable prototypes in the original feature space, enabling transparent interpretation of model behavior and adaptation. By formulating meta-learning as a cognitive-level analytical process, it provides a principled means to examine how fuzzy rules evolve across domains and tasks. Furthermore, a knowledge-guided regularization mechanism establishes a \emph{top-down--bottom-up integration}, where theoretical priors act as soft constraints while data-driven learning refines and extends them. Together, these contributions advance the methodological foundation of interpretable fuzzy systems by bridging symbolic reasoning and adaptive representation learning within a unified, extensible paradigm. Conceptually, the OR/OI meta-learning formulation introduced here can be extended to other cognitive domains, suggesting that the same rule-level adaptation principles could support transfer across tasks such as attention, working memory, or affective decoding.

\vspace*{-1em}  
\begin{small}
\bibliographystyle{IEEEtran}
\bibliography{ref}
\end{small}

\end{document}